\pdfoutput=1
\documentclass[runningheads,final]{llncs}

\makeatletter
\def\mdseries@tt{m}
\def\mdseries@rm{m}
\makeatother

\usepackage[T1]{fontenc}
\usepackage[utf8]{inputenc}

\usepackage[usenames,dvipsnames]{xcolor}
\definecolor{color0}{HTML}{404040}
\definecolor{color1}{HTML}{BD4E42}
\definecolor{color2}{HTML}{005B82}
\definecolor{color3}{HTML}{7D966E}
\definecolor{color4}{HTML}{D7AA50}

\usepackage{algorithmic}
\usepackage[nosumlimits]{amsmath}
\usepackage{array}
\usepackage[pdftex]{graphicx}
\usepackage{subcaption}
\usepackage{placeins}
\usepackage[obeyFinal]{todonotes}
\usepackage{url}
\usepackage{wrapfig}
\usepackage{xspace}
\usepackage{tikz}
\usetikzlibrary{arrows,arrows.meta,bending,positioning,calc,patterns,backgrounds,fit,decorations.pathreplacing,shapes.geometric}

\usepackage{ifdraft}
\usepackage{ifthen}

\usepackage{dblfloatfix}
\usepackage{paralist}
\usepackage{floatrow}
\usepackage{booktabs}

\usepackage[style=lncs,sorting=none,minbibnames=3]{biblatex}
\usepackage[normalem]{ulem}

\usepackage{caption}
\DeclareCaptionFont{white}{\color{white}}
\DeclareCaptionFormat{listing}{\colorbox{gray}{\parbox{\linewidth}{#1#2#3}}}
\usepackage{listings,lstautogobble}
\usepackage[newfloat,finalizecache=false,frozencache=true]{minted}
\setminted{fontsize=\footnotesize}
\setminted{breaklines=true}
\setminted{autogobble=true}
\setminted{fontfamily=courier}

\usepackage{hyperref}

\newcommand\myshade{85}
\colorlet{mylinkcolor}{violet}
\colorlet{mycitecolor}{YellowOrange}
\colorlet{myurlcolor}{Aquamarine}

\hypersetup{
  linkcolor  = mylinkcolor!\myshade!black,
  citecolor  = mycitecolor!\myshade!black,
  urlcolor   = myurlcolor!\myshade!black,
  colorlinks = true,
  breaklinks = true, %
}
\usepackage{breakurl} %

\usepackage{cleveref}
\Crefname{figure}{Fig.}{Figs.}
\crefname{figure}{fig.}{figs.}
\Crefname{section}{Sec.}{Secs.}
\crefname{section}{sec.}{secs.}
\Crefname{table}{Tab.}{Tabs.}
\crefname{table}{tab.}{tabs.}

\usepackage{xfrac}
\usepackage[detect-all,binary-units]{siunitx}
\DeclareSIUnit\operation{op}

\newcommand{\BSS}[1]{BSS\nobreakdash-#1}
\newcommand{\BrainScaleS}[1]{BrainScaleS\nobreakdash-#1}

\begin{document}

\title{\texttt{hxtorch}: PyTorch for BrainScaleS-2}
\subtitle{Perceptrons on Analog Neuromorphic Hardware}

\newcommand{\repeatthanks}{\textsuperscript{\thefootnote}}
\author{%
		Philipp Spilger\thanks{contributed equally.}
		\and
		Eric Müller\repeatthanks
		\and
		Arne Emmel
		\and
		Aron Leibfried
		\and
		Christian Mauch
		\and
		Christian Pehle
		\and
		Johannes Weis
		\and
		Oliver Breitwieser
		\and
		Sebastian Billaudelle
		\and
		Sebastian Schmitt
		\and
		Timo C.\ Wunderlich
		\and
		Yannik Stradmann
		\and\\
		Johannes Schemmel%
}

\authorrunning{P.\ Spilger \and E.\ Müller et al.}
\institute{Kirchhoff-Institute for Physics\\
Ruprecht-Karls-Universität Heidelberg, Germany\\
\email{\{pspilger,mueller\}@kip.uni-heidelberg.de}
}

\maketitle

\begin{abstract}
We present software facilitating the usage of the \mbox{\BrainScaleS{2}} analog neuromorphic hardware system
as an inference accelerator for artificial neural networks. %
The hardware is transparently integrated into the PyTorch machine learning framework using its extension interface. %
In particular, we support for vector-matrix multiplications and convolutions;
corresponding software-based autograd functionality is provided for hardware-in-the-loop training.
Automatic partitioning of neural networks onto one or multiple chips is supported.
We analyze implementation runtime overhead, provide measurements for existing setups and evaluate the results in terms of the hardware design limitations.
As an application of the introduced framework, we present a model that classifies activities of daily living with smartphone sensor data.


\end{abstract}

\section{Introduction}

Modern machine learning (ML) frameworks such as PyTorch or Tensorflow provide interfaces and tools to easily define and evaluate machine learning models \cite{he2019mlframeworks}.
They aim to reduce development effort and increase interoperability by providing a large collection of operators, algorithms, optimizers and analysis tools.
Using a high-level specification of the complete model, these packages construct a corresponding computational graph that is fundamentally agnostic to the substrate on which it is executed and optimized.
State-of-the-art machine learning frameworks support CPU and GPU-based backends that accelerate inference and training.
Compared to typical CPUs, GPUs are capable of processing data in a highly parallel fashion and are well-suited to the abundance of vector-matrix multiplications in artificial neural networks.

While GPUs represent a traditional approach to accelerating the execution of artificial neural networks, neuromorphic chips mimic biological neural networks and provide a research platform for computational neuroscience and beyond-Von-Neumann computation.
\BrainScaleS{2} is such a neuromorphic system; it includes analog neurons and can operate in both a spiking and non-spiking mode.
The non-spiking mode can be used to implement analog vector-matrix multiplication \cite{weis2020inference} while the spiking mode provides neuron dynamics accelerated relative to biology and CPU-based digital simulations \cite{wunderlich2019demonstrating_nourl}.
In this work, we present software that allows \BrainScaleS{2} to be used as a backend for PyTorch in the non-spiking mode (i.e.\ for vector-matrix multiplication).

\begin{wrapfigure}[11]{r}{0.33\textwidth}
	\vspace{-2\baselineskip} %
	\includegraphics[width=.527\textwidth]{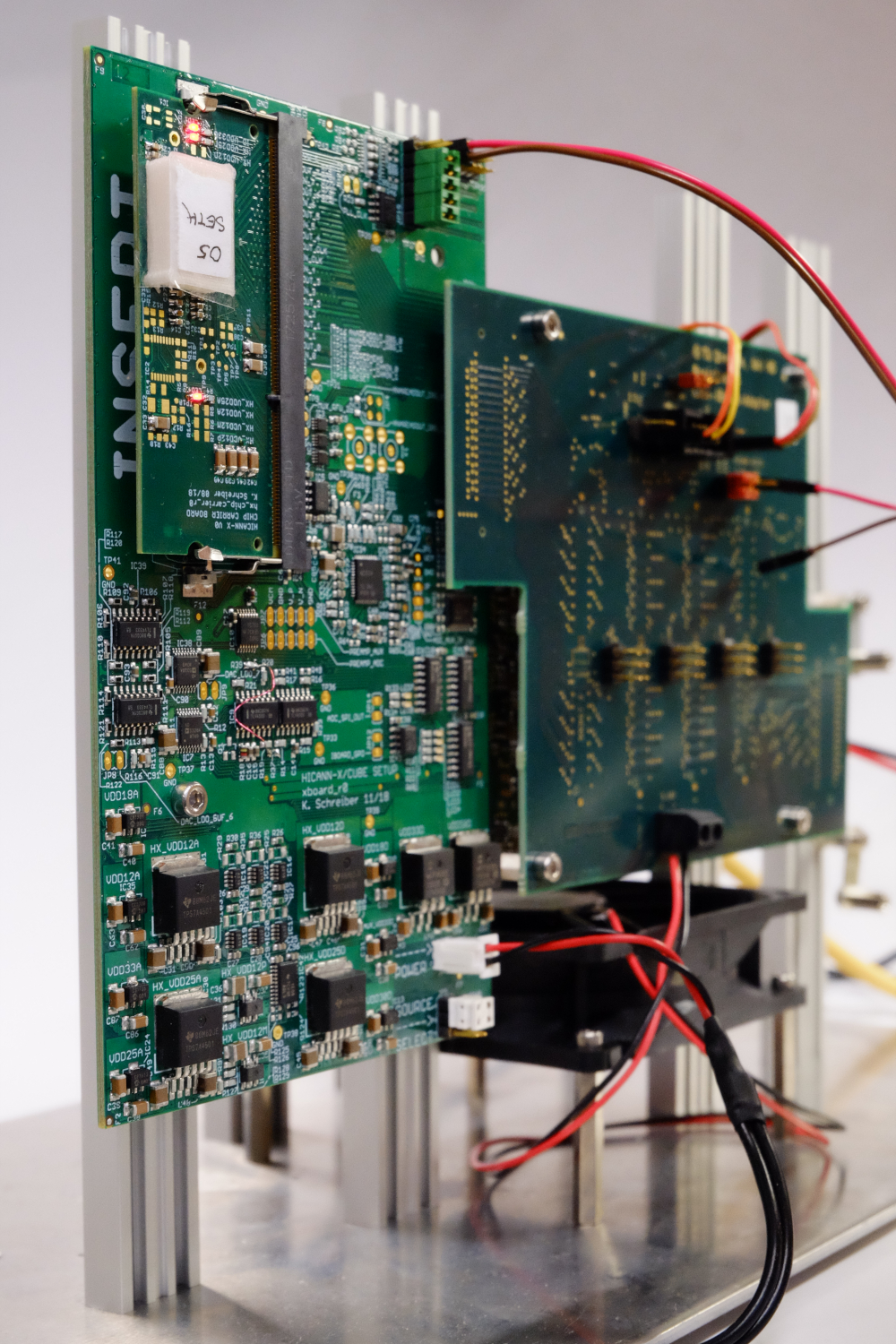}%
	\includegraphics[width=.473\textwidth]{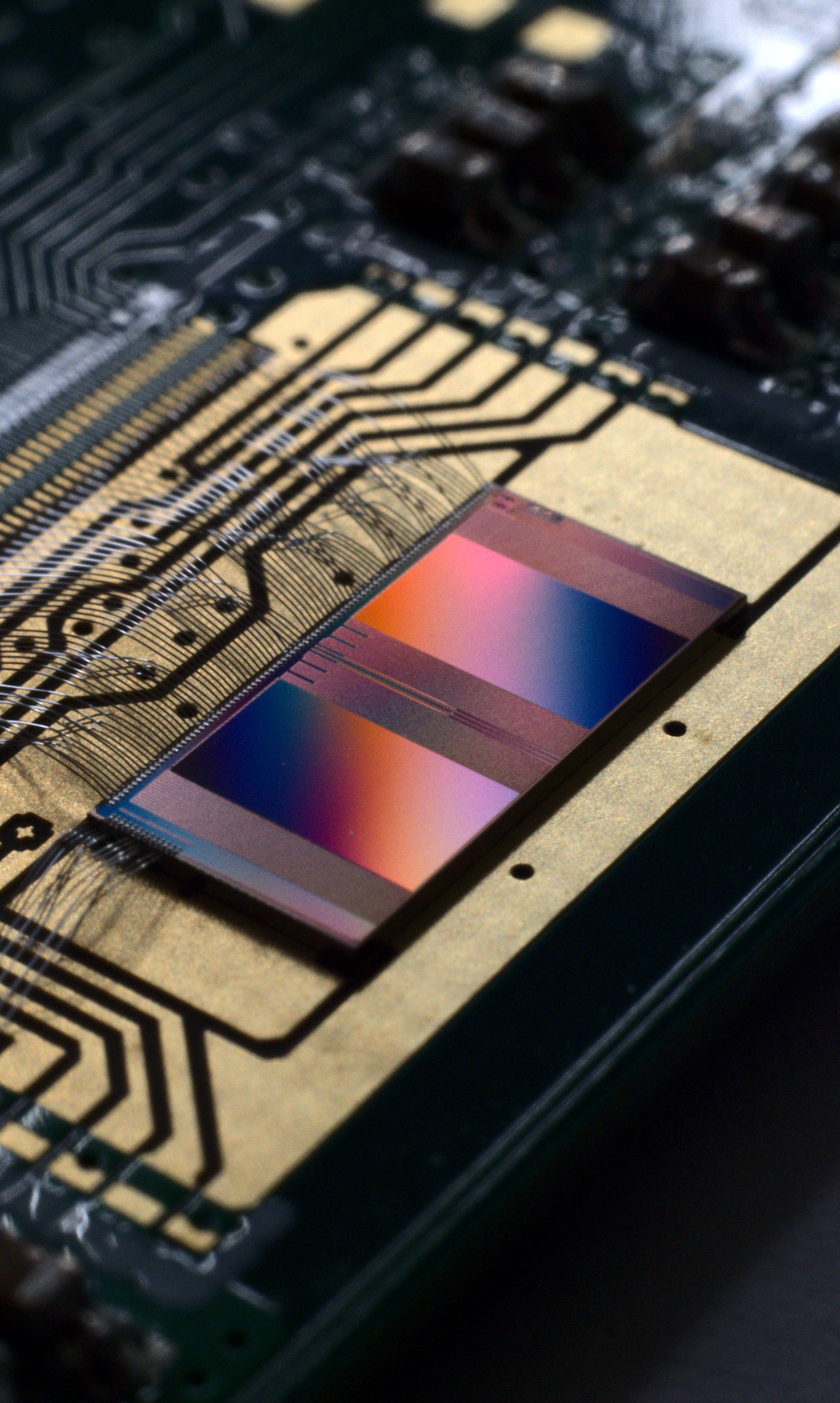}
	\vspace{-1.75\baselineskip}
	\caption{\label{fig:hicannx-setup}%
		\BSS{2} setup (left);
		a white dust cap covers the chip (right).
	}
\end{wrapfigure}%
\Cref{fig:hicannx-setup} shows the \BrainScaleS{2} (\BSS{2}) hardware setup providing \num{512} neurons per chip, see~\cite{schemmel2020accelerated_nourl}.
Synapses are arranged in a matrix above the neurons, containing \num{256} rows for inputs, and connecting to the neurons in columns.
Vector entries control how long a synapse row is activated.
The charge reaching the neurons is then the product of the vector entry and the matrix weight, as it is the product of time and current.
Neurons accumulate charge emitted by a column of synapses on the membrane capacitance, which completes the vector-matrix multiplication.
All neuron membrane voltages are digitized in parallel.
The resulting operation is
$x^\mathsf{T} W = y,$
where the inputs $x_i \in [0,31]$ are transposed multiplied by the weight matrix $W_{ij} \in [-63,63]$ for signed and $W_{ij} \in [0,63]$ for unsigned weights, resulting in outputs $y_j \in [-128,127]$.
The available resolution of vector entries therefore is 5 bits, synapse weights are 6 bits (plus a sign for signed weights), the result is 8 bits.
Two embedded SIMD microprocessors provide additional flexibility and support vector access to the synapse array.
In the current lab setup, host connectivity is provided by a 1GBit-Ethernet (GbE) link to an FPGA providing buffer memory and managing real-time access to the \BSS{2} chip.
Details on the hardware architecture can be found in \cite{weis2020inference}.

\section{Methods and Tools}\label{sec:methods_and_tools}

Simple and user-friendly interfaces are key components affecting the success of custom hardware accelerators.
Especially the PyTorch and Tensorflow machine learning frameworks are widely known and established in both, research and industry, cf.~\cite{he2019mlframeworks}.
Compared to Tensorflow, we experienced a simpler integration of the PyTorch build flow into our software environment:
the build flow is simpler and relies on standard \texttt{cmake},
and out-of-tree dependencies are tracked using standard mechanisms.
In conjunction with sufficient documentation, this is the reason we chose PyTorch~\cite{paszke2019pytorch_nourl_shorter} as a frontend for our accelerator hardware.
However, a similar integration into Tensorflow is possible. %

In terms of artificial neural networks, the \BSS{2} system provides support for a limited set of operations:
matrix multiplications and convolutions can be mapped onto analog multiplication and accumulation units.
Two embedded SIMD microprocessors can be used to perform additional operations.
However, the processors are not optimized for number crunching but rather serve as programmable controllers for the analog compute units.

The software we present in the next section builds upon the ``BrainScaleS Operating System'', a software stack providing multiple hardware access layers.
We make use of a register-level abstraction layer,
other components provide a reliable GbE-based communication channel to the hardware system
or integrate multiple hardware systems into the SLURM resource manager.
The authors provide an overview over the software components in~\cite{mueller2020bss1_nourl,mueller2020bss2ll_nourl}.

\section{Implementation}\label{sec:implementation}

In this section we give an overview of our software implementation providing a \BSS{2} hardware backend for PyTorch.
We start with the PyTorch integration and considerations regarding data and computational flow.
Afterwards, we describe our graph-based approach for describing and handling control and data flow on-chip as well as off-chip.
We partition operations into chunks fitting onto the available hardware and schedule hardware using just-in-time-based (JIT) execution.
Finally, we focus on hardware-specific aspects.

\subsection{PyTorch integration}
PyTorch provides an extensions interface which allows us to implement new custom operations without modifying the core source code or build system.
Integration of the computation executed on the accelerator hardware into PyTorch can be split into two parts.
Firstly, low-level operator primitives written in \texttt{C++} and wrapped to \texttt{Python} implementing the \texttt{torch.autograd.Function} interface provide means to execute the operation directly.
Since \BSS{2} is geared towards accelerating two-dimensional matrix multiplication, the low-level operator primitive implemented with direct hardware access is adhering to the interface of the \texttt{matmul} and \texttt{conv\{1,2,3\}d} operation.
Secondly, layers using the \texttt{torch.nn.Module} interface provide the ability to integrate the computation into an abstract model, a representation of the computation to be done without eagerly executing it.

\begin{figure}[tb]
	{%
		\resizebox{1.0\textwidth}{!}{\begin{tikzpicture}[node distance=0.4,
    every node/.style={fill=white, font=\sffamily}, align=center,>={Latex[width=2mm,length=2mm]},
    base/.style = {rectangle, draw=black,
                   minimum width=1cm, minimum height=0.5cm,
                   text centered, font=\sffamily},
    activityStarts/.style = {base, fill=color2!40},
    startstop/.style = {base, fill=color1!40},
    activityRuns/.style = {base, fill=color3!40},
    process/.style = {base, fill=color4!40,
                      font=\sffamily}]
	\node (pytorchin) [activityStarts] {PyTorch};
	\node (reshapein) [process, right=1.3 of pytorchin] {reshape};
	\node (transformin) [process, right=of reshapein] {convert};
	\node (partitioning) [process, right=of transformin] {partition};
	\node (execution) [process, right=of partitioning] {jit execute};
	\node (bss2) [startstop, above=of execution] {BSS-2};
	\node (transformout) [process, right=of execution] {convert};
	\node (reshapeout) [process, right=of transformout] {reshape};
	\node (pytorchout) [activityStarts, right=of reshapeout] {PyTorch};
	\node (matmultext) [above=of reshapeout, opacity=0.0, text opacity=1.0, xshift=0.14cm, yshift=0.15cm] {matmul};

	\draw[decorate,decoration={brace,amplitude=3pt,mirror}] (reshapein.south west)+(0,-0.1cm) -- ([yshift=-0.1cm]transformin.south east);
	\node (pre) [below=0.45cm of reshapein, anchor=center, xshift=0.85cm, opacity=0.0, text opacity=1.0] {preprocess};

	\draw[decorate,decoration={brace,amplitude=3pt,mirror}] (transformout.south west)+(0,-0.1cm) -- ([yshift=-0.1cm]reshapeout.south east);
	\node (post) [below=0.45cm of transformout, anchor=center, xshift=0.85cm, opacity=0.0, text opacity=1.0] {postprocess};

	\draw[->] (pytorchin) -- node[above=0.1cm, opacity=0, text opacity=1, align=center]{$\textsf{x}_{\textsf{float32}}$\\$\textsf{w}_{\textsf{float32}}$} (reshapein);
	\draw[->] (reshapein) -- (transformin);
	\draw[->] (transformin) -- node[above=0.3cm, fill=color0!20, opacity=0, text opacity=1]{$\textsf{x}_{\textsf{u5}},\ \textsf{w}_{\textsf{s6}^*}$} (partitioning);
	\draw[->] (partitioning) -- node[below=0.3cm, fill=color0!20, opacity=0, text opacity=1]{data flow graph} (execution);
	\draw[->, transform canvas={xshift=-0.05cm}] (execution.north) to[bend left=30] (bss2.south);
	\draw[->, transform canvas={xshift=0.05cm}] (bss2.south) to[bend left=30] (execution.north);
	\draw[->] (execution) -- node[above=0.3cm, fill=color0!20, opacity=0, text opacity=1]{$\textsf{y}_{\textsf{s8}}$} (transformout);
	\draw[->] (transformout) -- node[above=0.3cm,fill=color0!20, opacity=0, text opacity=1]{$\textsf{y}_{\textsf{float32}}$} (reshapeout);
	\draw[->] (reshapeout) -- (pytorchout);

	\begin{scope}[on background layer]
		\node[draw, color=color0!20, fit=(reshapein) (transformin) (partitioning) (execution) (bss2) (transformout) (reshapeout) (pre) (post)] (matmulbox) {};
	\end{scope}
\end{tikzpicture}}
	}
	{%
		\vspace{-1\baselineskip}
		\caption{\label{fig_matmul}%
		Implementation of the \texttt{matmul} operation executed on \BSS{2}.
		Inputs and weights of type \texttt{float32} are reshaped such that the input batch space is one dimensional;
		types are converted to $5$-bit unsigned inputs and $6$-bit ``signed'' weights ($[-63,63]$).
		A partitioner performs splitting and placement of the weight matrix onto the hardware resources.
		The resulting data flow graph is then used for JIT execution, which constructs an instruction stream sent to the hardware, see~\cite{weis2020inference}, and decodes the result data received from the hardware using the software architecture described in \cite{mueller2020bss2ll_nourl}.
		The resulting digitized $8$-bit neuron membrane potential values are converted to \texttt{float32}, reshaped to match the originally provided type/shape from PyTorch and returned as the result of the operation.
		}
	}
\end{figure}
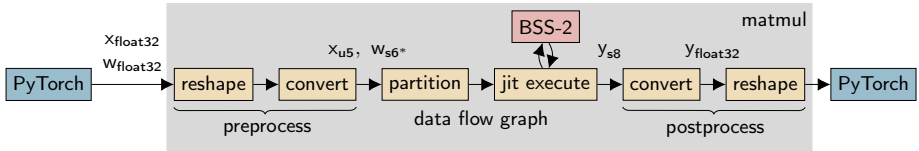

\paragraph{Forward pass}
The forward pass of the \texttt{matmul} operation is comprised of a linear sequence of transformations to and from the hardware.
\Cref{fig_matmul} shows its internal implementation.
Tensors from PyTorch are preprocessed to match types and shape of the hardware.
A partitioner then places the operation onto available hardware resources, cf.~\cref{section_partitioning}.
The resulting custom data flow graph is then traversed by JIT execution, cf.~\cref{section_graph}.
Measured activations are postprocessed to resemble the expected shape and type and returned.

\paragraph{Backward pass --- hardware in-the-loop}
PyTorch features intrinsic support for automatic differentiation of a sequence of operations via the \texttt{torch.autograd.\allowbreak{}Function} interface, which is used for the low-level operations on \BSS{2}.
Supply of a \texttt{backward()} function allows integration of custom operations into this framework.
The output gradient as well as saved state of the forward pass of the operation is used to compute a gradient for the inputs.
The analog accelerator's usability for computing the gradient of an operation is limited due to fixed-pattern and temporal noise.
However using the results obtained from the hardware execution of the forward pass together with a software model to calculate the gradients allows mitigation of hardware distortions and resolution limitations during training.
This method is adapted from~\cite{schmitt2017hwitl_nourl}, where a combination of a software model and the forward pass data from \BSS{1} hardware runs are used to train a spiking network, thereby training with \textit{hardware in the loop}.
This approach is also suited for spike-based modeling, e.g., \cite{cramer2020training_nourl}. %
The software model employed here is much simpler in that it only resembles the backward pass of the conventional \texttt{matmul} operation.
The assumption is that the direction of the gradient is correct and potential scaling between the software model and the hardware execution will be mitigated by adjusting the learning rate.

\begin{figure}[tb]
		\newcommand{\kernel}[2]{
	\draw[draw=none, fill=color4!80] (#1,#2) rectangle ++(1,1);
	\draw[draw=none, fill=color4!60] (#1 - 0.05,#2 + 0.05) rectangle ++(1,1);
	\draw[draw=none, fill=color4!40] (#1 - 0.1,#2 + 0.1) rectangle ++(1,1);

	\draw[draw] (#1 - 0.1,#2 + 0.1) rectangle ++(0.5,0.5);
	\draw[draw] (#1 - 0.1 + 0.5,#2 + 0.1) rectangle ++(0.5,0.5);
	\draw[draw] (#1 - 0.1,#2 + 0.1 + 0.5) rectangle ++(0.5,0.5);
	\draw[draw] (#1 - 0.1 + 0.5,#2 + 0.1 + 0.5) rectangle ++(0.5,0.5);

	\node (k) at (#1 -0.1 + 0.25, #2 + 0.1 + 0.5 + 0.25) {$\text{k}_{00}$};
	\node (k) at (#1 -0.1 + 0.25, #2 + 0.1 + 0.25) {$\text{k}_{10}$};
	\node (k) at (#1 -0.1 + 0.25 + 0.5, #2 + 0.1 + 0.5 + 0.25) {$\text{k}_{01}$};
	\node (k) at (#1 -0.1 + 0.25 + 0.5, #2 + 0.1 + 0.25) {$\text{k}_{11}$};
}

\newcommand{\matmul}[2]{
	\foreach \y in {0,...,3} {
		\draw[draw, fill=color0!20] (#1,#2 + \y * 0.4) rectangle ++(0.5,0.4);
	}

	\node (x) at (#1 + 0.25, #2 + 0.2) {$\text{x}_{11}$};
	\node (x) at (#1 + 0.25, #2 + 0.2 + 0.4) {$\text{x}_{10}$};
	\node (x) at (#1 + 0.25, #2 + 0.8 + 0.2) {$\text{x}_{01}$};
	\node (x) at (#1 + 0.25, #2 + 0.2 + 1.2) {$\text{x}_{00}$};

	\foreach \y in {0,...,3} {
		\draw[draw, fill=color4!40] (1 + #1 + 0 * 0.5,#2 + \y * 0.4) rectangle ++(0.5,0.4);
	}
	\node (k) at (#1 + 0.25 + 1 + 0 * 0.5, #2 + 0.2) {$\text{k}_{11}$};
	\node (k) at (#1 + 0.25 + 1 + 0 * 0.5, #2 + 0.2 + 0.4) {$\text{k}_{10}$};
	\node (k) at (#1 + 0.25 + 1 + 0 * 0.5, #2 + 0.8 + 0.2) {$\text{k}_{01}$};
	\node (k) at (#1 + 0.25 + 1 + 0 * 0.5, #2 + 0.2 + 1.2) {$\text{k}_{00}$};

	\foreach \y in {0,...,3} {
		\draw[draw, fill=color4!60] (1 + #1 + 1 * 0.5,#2 + \y * 0.4) rectangle ++(0.5,0.4);
	}
	\node (k) at (#1 + 0.25 + 1 + 1 * 0.5, #2 + 0.2) {$\text{k}_{11}$};
	\node (k) at (#1 + 0.25 + 1 + 1 * 0.5, #2 + 0.2 + 0.4) {$\text{k}_{10}$};
	\node (k) at (#1 + 0.25 + 1 + 1 * 0.5, #2 + 0.8 + 0.2) {$\text{k}_{01}$};
	\node (k) at (#1 + 0.25 + 1 + 1 * 0.5, #2 + 0.2 + 1.2) {$\text{k}_{00}$};

	\foreach \y in {0,...,3} {
		\draw[draw, fill=color4!80] (1 + #1 + 2 * 0.5,#2 + \y * 0.4) rectangle ++(0.5,0.4);
	}
	\node (k) at (#1 + 0.25 + 1 + 2 * 0.5, #2 + 0.2) {$\text{k}_{11}$};
	\node (k) at (#1 + 0.25 + 1 + 2 * 0.5, #2 + 0.2 + 0.4) {$\text{k}_{10}$};
	\node (k) at (#1 + 0.25 + 1 + 2 * 0.5, #2 + 0.8 + 0.2) {$\text{k}_{01}$};
	\node (k) at (#1 + 0.25 + 1 + 2 * 0.5, #2 + 0.2 + 1.2) {$\text{k}_{00}$};

	\foreach \x in {0,...,2} {
		\draw[draw, fill=color0!20] (1 + #1 + \x * 0.5,#2 - 0.8) rectangle ++(0.5,0.4);
	}
	\node (y) at (#1 + 0.25 + 1, #2 + 0.2 - 0.8) {$\text{y}_{00}$};
	\node (y) at (#1 + 0.25 + 1 + 0.5, #2 + 0.2 - 0.8) {$\text{y}_{00}$};
	\node (y) at (#1 + 0.25 + 1 + 2 * 0.5, #2 + 0.2 - 0.8) {$\text{y}_{00}$};

	\draw[->] (#1 + 0.55, #2 + 0.8) -- (#1 + 0.95, #2 + 0.8);
	\draw[->] (#1 + 1.75, #2 - 0.05) -- (#1 + 1.75, #2 - 0.35);
}

\newcommand{\movedmatmul}[2]{
	\foreach \y in {0,...,3} {
		\draw[draw, fill=color0!20] (#1,#2 + \y * 0.4) rectangle ++(0.5,0.4);
	}

	\node (x) at (#1 + 0.25, #2 + 0.2) {$\text{x}_{1\textcolor{color1}2}$};
	\node (x) at (#1 + 0.25, #2 + 0.2 + 0.4) {$\text{x}_{1\textcolor{color1}1}$};
	\node (x) at (#1 + 0.25, #2 + 0.8 + 0.2) {$\text{x}_{0\textcolor{color1}2}$};
	\node (x) at (#1 + 0.25, #2 + 0.2 + 1.2) {$\text{x}_{0\textcolor{color1}1}$};

	\foreach \y in {0,...,3} {
		\draw[draw, fill=color4!40] (1 + #1 + 0 * 0.5,#2 + \y * 0.4) rectangle ++(0.5,0.4);
	}
	\node (k) at (#1 + 0.25 + 1 + 0 * 0.5, #2 + 0.2) {$\text{k}_{11}$};
	\node (k) at (#1 + 0.25 + 1 + 0 * 0.5, #2 + 0.2 + 0.4) {$\text{k}_{10}$};
	\node (k) at (#1 + 0.25 + 1 + 0 * 0.5, #2 + 0.8 + 0.2) {$\text{k}_{01}$};
	\node (k) at (#1 + 0.25 + 1 + 0 * 0.5, #2 + 0.2 + 1.2) {$\text{k}_{00}$};

	\foreach \y in {0,...,3} {
		\draw[draw, fill=color4!60] (1 + #1 + 1 * 0.5,#2 + \y * 0.4) rectangle ++(0.5,0.4);
	}
	\node (k) at (#1 + 0.25 + 1 + 1 * 0.5, #2 + 0.2) {$\text{k}_{11}$};
	\node (k) at (#1 + 0.25 + 1 + 1 * 0.5, #2 + 0.2 + 0.4) {$\text{k}_{10}$};
	\node (k) at (#1 + 0.25 + 1 + 1 * 0.5, #2 + 0.8 + 0.2) {$\text{k}_{01}$};
	\node (k) at (#1 + 0.25 + 1 + 1 * 0.5, #2 + 0.2 + 1.2) {$\text{k}_{00}$};

	\foreach \y in {0,...,3} {
		\draw[draw, fill=color4!80] (1 + #1 + 2 * 0.5,#2 + \y * 0.4) rectangle ++(0.5,0.4);
	}
	\node (k) at (#1 + 0.25 + 1 + 2 * 0.5, #2 + 0.2) {$\text{k}_{11}$};
	\node (k) at (#1 + 0.25 + 1 + 2 * 0.5, #2 + 0.2 + 0.4) {$\text{k}_{10}$};
	\node (k) at (#1 + 0.25 + 1 + 2 * 0.5, #2 + 0.8 + 0.2) {$\text{k}_{01}$};
	\node (k) at (#1 + 0.25 + 1 + 2 * 0.5, #2 + 0.2 + 1.2) {$\text{k}_{00}$};

	\foreach \x in {0,...,2} {
		\draw[draw, fill=color0!20] (1 + #1 + \x * 0.5,#2 - 0.8) rectangle ++(0.5,0.4);
	}
	\node (y) at (#1 + 0.25 + 1, #2 + 0.2 - 0.8) {$\text{y}_{0\textcolor{color1}1}$};
	\node (y) at (#1 + 0.25 + 1 + 0.5, #2 + 0.2 - 0.8) {$\text{y}_{0\textcolor{color1}1}$};
	\node (y) at (#1 + 0.25 + 1 + 2 * 0.5, #2 + 0.2 - 0.8) {$\text{y}_{0\textcolor{color1}1}$};

	\draw[->] (#1 + 0.55, #2 + 0.8) -- (#1 + 0.95, #2 + 0.8);
	\draw[->] (#1 + 1.75, #2 - 0.05) -- (#1 + 1.75, #2 - 0.35);
}

\begin{tikzpicture}[node distance=1.2cm,
    every node/.style={font=\sffamily},>={Latex[width=2mm,length=2mm]},
    fill upper half/.style={path picture={\fill[#1] (path picture bounding box.north west) rectangle (path picture bounding box.east);}},
    fill upper left/.style={path picture={\fill[#1] (path picture bounding box.north west) rectangle (path picture bounding box);}},
    fill left half/.style={path picture={\fill[#1] (path picture bounding box.north west) rectangle (path picture bounding box.south);}}]

	\draw[draw=none, fill=color0!20] (0,-0.2) rectangle ++(2,1.2);
	\kernel{0}{0}
	\draw[draw] (1.5,0.5) rectangle ++(0.5,0.5);
	\node (x) at (1.5 + 0.25, 0.5 + 0.25) {$\text{x}_{ij}$};

	\draw[draw=none, fill=color0!20] (6.5,-0.2) rectangle ++(2,1.2);
	\draw[draw=none, fill=color4!30] (6.5,0) rectangle ++(1,1);
	\kernel{7}{0}

	\matmul{2.5}{0}

	\movedmatmul{9}{0}
	\draw[->, color=color1] (6.5 - 0.1, 0.6) -- ++(0.5,0);
	\node at (2.25, -0.6) {(a)};
	\node at (2.25 + 6.5, -0.6) {(b)};
\end{tikzpicture}
		\vspace{-.5\baselineskip}
		\caption{\label{fig_conv}%
		Transformation of a 2-d convolution ((a), (b) left) of inputs $x_{ij}$ with kernel $k_{ij}$, three channels denoted by different shades and a stride of $1$ (moved from (a) to (b)) to a multiplication.
		The resulting matrix ((a), (b) right) is constant for all kernel positions, which is efficient in terms of reconfiguration of the weights while leading to overlapping inputs for different kernel positions.
		}
\end{figure}
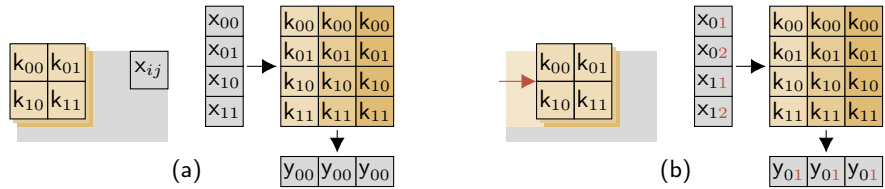

\paragraph{Mapping convolutions}
In order to support convolutions, a \texttt{convNd} operation is implemented using the \texttt{matmul} operation.
The convolution is transformed to a matrix multiplication by unrolling the kernel into the vertical dimension, placing all kernel channels horizontally aside each other and traversing the input such that each operation is equivalent to the original kernel applied at a certain position.
\Cref{fig_conv} shows the transformation exemplary for a \texttt{conv2d} operation.
The transformations are implemented using the \texttt{C++}-API of PyTorch, which allows most operations to be in-place modifications of tensor shapes only.

\subsection{Graph Representation and Just-in-time Execution}%
\label{section_graph}%
\label{section_execution}%
A major task of neural network accelerator operation is the tracking of data flow to and from the host as well as on the hardware substrate itself.
In the case of \BSS{2} and in addition to general input and output properties, heterogeneous entities, e.g., synapses and neuron circuits, need to be configured.
Limited hardware resources require a temporal reuse of hardware substrate to compute a larger operation over the course of multiple inter-dependent executions.

\begin{figure}[tb]
\centering
\newcommand{\syndrvsingle}[3]{
	\draw[draw, fill=#3] (#1, #2)--(#1,#2 + 0.3)--(#1 + 0.3, #2 + 0.15)--(#1,#2);
	\draw[->] (#1 + 0.35, #2 + 0.145) -- (#1 + 0.9, #2 + 0.15);
}
\newcommand{\syndrv}[3]{
	\syndrvsingle{#1}{#2}{#3}
	\syndrvsingle{#1 + 0.05}{#2 - 0.1}{#3}

}
\newcommand{\mac}[3]{
	\draw[draw, fill=color0!20, opacity=#3] (#1, #2 - 0.75) rectangle (#1 + 6.55, #2 + 3);
	\node[draw=none, anchor=west, align=left, opacity=#3] (se) at (#1 + 0.1, #2 - 0.4 + 0.1) {execution instance};

	\draw[draw, fill=color2!20, opacity=#3] (#1 + 3, #2 + 1) rectangle (#1 + 3 + 1.8, #2 + 1 + 1.8);
	\node[draw=none, align=center, anchor=center, opacity=#3] at (#1 + 3 + 0.9, #2 + 1 + 0.9) {synapse \\ matrix};

	\syndrv{#1 + 2.0}{#2 + 2.5}{color2!60}
	\syndrv{#1 + 2.0}{#2 + 2.033}{color2!60}
	\syndrv{#1 + 2.0}{#2 + 1.567}{color2!60}
	\syndrv{#1 + 2.0}{#2 + 1.1}{color2!60}

	\node[draw, fill=color3!40, anchor=west, align=center, opacity=#3] (load) at (#1 + 0.25, #2 + 1.95) {external \\ load};

	\draw[->, opacity=#3] (load) -- (#1 + 1.95, #2 + 2.6);
	\draw[->, opacity=#3] (load) -- (#1 + 1.95, #2 + 2.133);
	\draw[->, opacity=#3] (load) -- (#1 + 1.95, #2 + 1.667);
	\draw[->, opacity=#3] (load) -- (#1 + 1.95, #2 + 1.2);

	\node[draw, rectangle, fill=color4!20, opacity=#3, minimum width=1.8cm, align=center, anchor=center] (neurons) at (#1 + 3 + 0.9, #2 + 0.3 + 0.15) {neurons};

	\node[draw, rectangle, fill=color1!20, opacity=#3, minimum width=1.8cm, inner sep=1.5, align=center, anchor=center] (adc) at (#1 + 3 + 0.9, #2 - 0.4 + 0.15) {digitize};

	\draw[->, opacity=#3] (#1 + 3.9, #2 + 0.95) -- (neurons);
	\draw[->, opacity=#3] (neurons) -- (adc);

	\node[draw, fill=color1!40, opacity=#3, anchor=east, align=center] (store) at (#1 + 3 + 1.8 + 1.5, #2 - 0.4 + 0.15) {store};
	\draw[->, opacity=#3] (adc) -- (store);
}

\begin{tikzpicture}[node distance=1.2cm,
    every node/.style={font=\sffamily},>={Latex[width=2mm,length=2mm]},
    square/.style={regular polygon,regular polygon sides=4},
    fill upper half/.style={path picture={\fill[#1] (path picture bounding box.north west) rectangle (path picture bounding box.east);}},
    fill upper left/.style={path picture={\fill[#1] (path picture bounding box.north west) rectangle (path picture bounding box);}},
    fill left half/.style={path picture={\fill[#1] (path picture bounding box.north west) rectangle (path picture bounding box.south);}}]

	\draw[draw, fill=color0!20] (7, 0) rectangle (8 + 4, 0 + 2);

	\node[draw=none, anchor=north east] (se) at (7 + 4.95, 0 + 1.95) {execution instance};
	\node[draw, circle, fill=color4!60] (+) at (7 + 2.5, 0 + 1) {+};
	\node[draw, fill=color3!40, anchor=west] (load1) at (7 + 0.25, 0 + 0.5) {load};
	\node[draw, fill=color3!40, anchor=west] (load2) at (7 + 0.25, 0 + 1.5) {load};
	\node[draw, fill=color1!40, anchor=east, align=center] (store) at (7 + 4.75, 0 + 1) {external \\ store};

	\draw[->] (load1) -- (+);
	\draw[->] (load2) -- (+);
	\draw[->] (+) -- (store);

	\mac{0.1}{0.75}{0.3}
	\mac{0}{1.75}{1.0}

	\draw[->, opacity=0.3] (3 + 1.8 + 1.5 + 0.1, 0.5) -- (load1);
	\draw[->] (3 + 1.8 + 1.5, 1.5) -- (load2);

\end{tikzpicture}
\vspace{-1.5\baselineskip}
\caption{
Graph representation of two matrix multiplications (left) followed by an addition (right).
Each multiplication as well as the addition are separately executed.
The data flow between execution instances (gray boxes) comprises stores and loads of measured neuron activations.
Vertices within the individual execution instances represent the on-chip data flow and hardware configuration.
}
\label{fig_graph}
\end{figure}
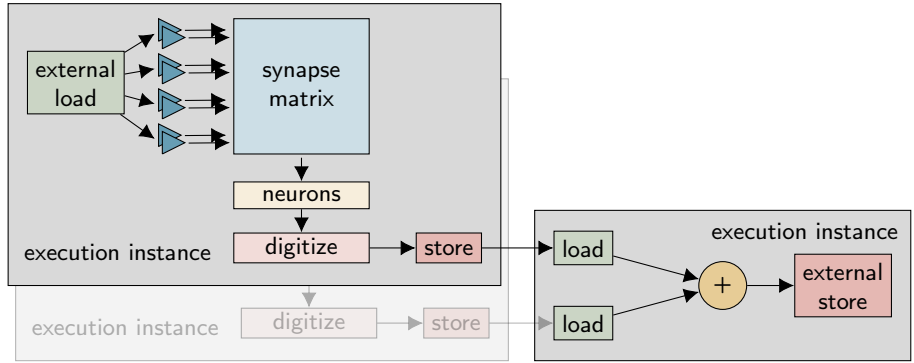

These demands are met by a hardware-centric data flow graph.
It describes the on-chip data flow as well as input and output.
While the former is used for hardware configuration,
the latter links individual execution instances in a \emph{dependency graph}.
Vertices represent statically configurable computation or hardware circuits.
A heterogeneous set of edge types allows to express analog and digital signal/data flow in a unified interface.
Support of batched input data mitigates comparably long static configuration times.
A static-single-assignment builder pattern facilitates correct configuration.
\Cref{fig_graph} shows an exemplary graph. %

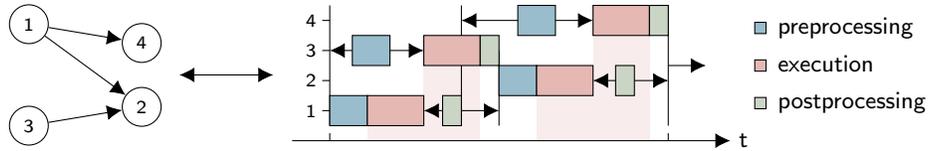
\begin{figure}[tb]
\centering
\newcommand{\legendentry}[4]{
	\draw[draw, fill=#4] (#1, #2) rectangle (#1 + 0.15, #2 + 0.15);
	\node[inner sep=0, anchor=base west] at (#1 + 0.3, #2) {#3};
}

\begin{tikzpicture}[node distance=1.2cm,
    every node/.style={font=\sffamily},>={Latex[width=2mm,length=2mm]},
    fill upper half/.style={path picture={\fill[#1] (path picture bounding box.north west) rectangle (path picture bounding box.east);}},
    fill upper left/.style={path picture={\fill[#1] (path picture bounding box.north west) rectangle (path picture bounding box);}},
    fill left half/.style={path picture={\fill[#1] (path picture bounding box.north west) rectangle (path picture bounding box.south);}}]

	\node[draw, circle] (e1) at (0, 1.55) {\scriptsize{1}};
	\node[draw, circle] (e3) at (0, 0.2) {\scriptsize{3}};
	\node[draw, circle] (e2) at (1.5, 0.45) {\scriptsize{2}};
	\node[draw, circle] (e4) at (1.5, 1.3) {\scriptsize{4}};

	\draw[->] (e1) -- (e4);
	\draw[->] (e1) -- (e2);
	\draw[->] (e3) -- (e2);

	\draw[draw=none, fill=color1!10] (4.5, 0) rectangle (5.25, 0.2);
	\draw[draw=none, fill=color1!10] (5.25, 0) rectangle (6, 1.2);
	\draw[draw=none, fill=color1!10] (6.75, 0) rectangle (7.5, 0.7);
	\draw[draw=none, fill=color1!10] (7.5, 0) rectangle (8.25, 1.7);

	\node (te) at (9.5, 0) {t};

	\draw[->] (3.5, 0) -- (te);

	\draw[<->] (2, 0.875) -- ++(1.25, 0);

	\draw[] (4, 0.2) -- (4, 1.8);
	\draw[->] (8.5, 1) -- ++(0.5, 0);

	\draw[] (5.75, 0.2) -- (5.75, 1.8);
	\draw[] (6.25, 0.2) -- (6.25, 1.2);
	\draw[] (8.5, 0.2) -- (8.5, 1.8);

	\draw[draw, fill=color2!40] (4, 0.2) rectangle ++(0.5, 0.4);
	\draw[draw, fill=color1!40] (4.5, 0.2) rectangle ++(0.75, 0.4);
	\draw[draw, fill=color3!40] (5.5, 0.2) rectangle ++(0.25, 0.4);

	\draw[draw, fill=color2!40] (4.3, 1.0) rectangle ++(0.5, 0.4);
	\draw[draw, fill=color1!40] (5.25, 1.0) rectangle ++(0.75, 0.4);
	\draw[draw, fill=color3!40] (6, 1.0) rectangle ++(0.25, 0.4);

	\draw[draw, fill=color2!40] (6.25, 0.6) rectangle ++(0.5, 0.4);
	\draw[draw, fill=color1!40] (6.75, 0.6) rectangle ++(0.75, 0.4);
	\draw[draw, fill=color3!40] (7.8, 0.6) rectangle ++(0.25, 0.4);

	\draw[draw, fill=color2!40] (6.5, 1.4) rectangle ++(0.5, 0.4);
	\draw[draw, fill=color1!40] (7.5, 1.4) rectangle ++(0.75, 0.4);
	\draw[draw, fill=color3!40] (8.25, 1.4) rectangle ++(0.25, 0.4);

	\draw[<-] (4, 1.2) -- ++(0.3, 0);
	\draw[<-] (5.25, 0.4) -- ++(0.25, 0);
	\draw[<-] (5.75, 1.6) -- ++(0.75, 0);
	\draw[<-] (7.5, 0.8) -- ++(0.3, 0);

	\draw[->] (4.8, 1.2) -- ++(0.45, 0);
	\draw[->] (5.75, 0.4) -- ++(0.5, 0);
	\draw[->] (7, 1.6) -- ++(0.5, 0);
	\draw[->] (8.05, 0.8) -- ++(0.45, 0);

	\node at (3.75, 0.4) {\scriptsize{1}};
	\node at (3.75, 0.8) {\scriptsize{2}};
	\node at (3.75, 1.2) {\scriptsize{3}};
	\node at (3.75, 1.6) {\scriptsize{4}};
	\draw[] (3.9, 0.4) -- ++(0.1, 0);
	\draw[] (3.9, 0.8) -- ++(0.1, 0);
	\draw[] (3.9, 1.2) -- ++(0.1, 0);
	\draw[] (3.9, 1.6) -- ++(0.1, 0);

	\draw[] (4, 0) -- (4, 0.1);
	\draw[] (8.5, 0) -- (8.5, 0.1);

	\legendentry{9.65}{0.925}{execution}{color1!40}
	\legendentry{9.65}{1.425}{preprocessing}{color2!40}
	\legendentry{9.65}{0.425}{postprocessing}{color3!40}

\end{tikzpicture}
\vspace{-1.5\baselineskip}
\caption{%
Just-in-time execution (right) of a dependency graph (left) consisting of four execution instances.
Four runs are scheduled onto the accelerator hardware.
Pre- and postprocessing of independent execution instances can overlap in time.
}
\label{fig_execution}
\end{figure}

Individual execution instances in the dependency graph are executed just in time.
Every instance can be split into a sequence of tasks:
preprocessing,
build of the instruction stream,
execution on the accelerator hardware
and postprocessing of the received results.
Assuming that concurrency on the host computer is not a limiting factor, pre/postprocessing of non-sequentially-dependent execution instances can be parallelized leading to a saturation of the accelerator hardware usage.
\Cref{fig_execution} exemplifies an dependency graph and a possible execution flow.

\subsection{Partitioning}\label{section_partitioning}
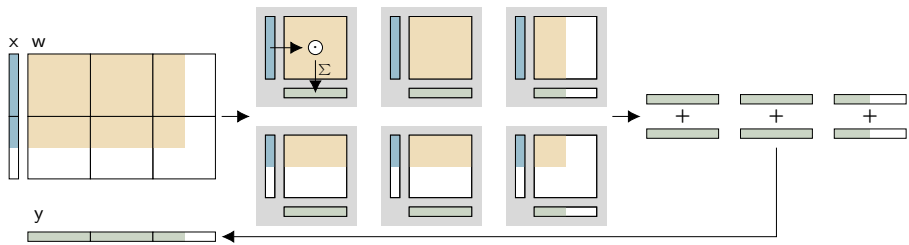
\begin{figure}[tb]
		\resizebox{\textwidth}{!}{\newcommand{\singleexecution}[5]{
	\draw[on background layer, draw=none, fill=color0!20] (-0.15 + #1, #2 - 1 -0.45) rectangle ++(1.6,1.6);
	\draw[draw, fill=white] (#1,#2) rectangle ++(0.15,-1);
	\draw[draw, #3=color2!40] (#1,#2) rectangle ++(0.15,-1);
	\draw[draw, fill=white] (0.3 + #1,#2) rectangle ++(1,-1);
	\draw[draw, #4=color4!40] (0.3 + #1,#2) rectangle ++(1,-1);
	\draw[draw, fill=white] (0.3 + #1,#2 - 1 -0.3) rectangle ++(1,0.15);
	\draw[draw, #5=color3!40] (0.3 + #1,#2 - 1 -0.3) rectangle ++(1,0.15);
}

\begin{tikzpicture}[node distance=1.2cm,
    every node/.style={font=\sffamily},>={Latex[width=2mm,length=2mm]},
    fill upper half/.style={path picture={\fill[#1] (path picture bounding box.north west) rectangle (path picture bounding box.east);}},
    fill upper left/.style={path picture={\fill[#1] (path picture bounding box.north west) rectangle (path picture bounding box);}},
    fill left half/.style={path picture={\fill[#1] (path picture bounding box.north west) rectangle (path picture bounding box.south);}}]

	\draw[draw, fill=color2!40] (0,1) rectangle ++(0.15,-1);
	\draw[draw, fill upper half=color2!40] (0,0) rectangle ++(0.15,-1);
	\node (x) at (0.075,1+0.175) {x};

	\draw[draw, fill=color4!40] (0.3 + 1,1) rectangle ++(1,-1);
	\draw[draw, fill=color4!40] (0.3,1) rectangle ++(1,-1);
	\draw[draw, fill upper half=color4!40] (0.3,0) rectangle ++(1,-1);
	\draw[draw, fill upper half=color4!40] (0.3 + 1,0) rectangle ++(1,-1);
	\draw[draw, fill left half=color4!40] (2.3,1) rectangle ++(1,-1);
	\draw[draw, fill upper left=color4!40] (2.3,0) rectangle ++(1,-1);
	\node (w) at (0.3 + 0.175,1+.175) {w};

	\draw[draw, fill=color3!40] (0.3,-2) rectangle ++(1,0.15);
	\draw[draw, fill=color3!40] (1.3,-2) rectangle ++(1,0.15);
	\draw[draw, fill left half=color3!40] (2.3,-2) rectangle ++(1,0.15);
	\node[anchor=south] (y) at (0.3 + 0.175,-2+0.175) {y};

	\singleexecution{4.1}{1.6}{fill}{fill}{fill}
	\singleexecution{4.1}{-0.3}{fill upper half}{fill upper half}{fill}

	\singleexecution{6.1}{1.6}{fill}{fill}{fill}
	\singleexecution{6.1}{-0.3}{fill upper half}{fill upper half}{fill}

	\singleexecution{8.1}{1.6}{fill}{fill left half}{fill left half}
	\singleexecution{8.1}{-0.3}{fill upper half}{fill upper left}{fill left half}

	\draw[draw, fill=color3!40] (9.7 + 0.5,0.2) rectangle ++(1.15,0.15);
	\draw[draw, fill=color3!40] (9.7 + 0.5,-0.35) rectangle ++(1.15,0.15);
	\draw[draw, fill=color3!40] (11.2 + 0.5,0.2) rectangle ++(1.15,0.15);
	\draw[draw, fill=color3!40] (11.2 + 0.5,-0.35) rectangle ++(1.15,0.15);
	\draw[draw, fill left half=color3!40] (12.7 + 0.5,0.2) rectangle ++(1.15,0.15);
	\draw[draw, fill left half=color3!40] (12.7 + 0.5,-0.35) rectangle ++(1.15,0.15);
	\node (+) at (10.575 + 0.2,0.0) {+};
	\node (+) at (12.075 + 0.2,0.0) {+};
	\node (+) at (13.575 + 0.2,0.0) {+};

	\draw[->] (3.4, 0.0) -- (3.85, 0.0);
	\draw[<-,to path={-| (\tikztotarget)}] (3.4, -1.925) -| (12.275, -0.5);
	\draw[->] (9.65, 0) -- (10.1, 0);

	\draw[->] (4.175, 1.1) -- ++(0.525, 0.0);
	\node[draw, circle, fill=white, inner sep=0.6, minimum size=0.3, text centered, anchor=mid] (m) at (4.9, 1.1) {$\cdot$};
	\draw[->] (4.9, 0.9) -- ++(0.0, -0.525);
	\node (+) at (5.05, 0.75) {\tiny$\sum$};

\end{tikzpicture}}
		\caption{\label{fig_partitioning}%
		Partitioning an operation too large to fit on a single synapse array.
		Top left: the input $x$ is multiplied with the weight matrix $w$.
		Inputs and weights are split at the black boundaries representing the shape of a hardware synapse array.
		Middle: as split operations are independent, they are allocated and executed individually.
		Right: split results in the row dimension are summed digitally,
		results in the column dimension are concatenated leading to the result $y$ (bottom left) of the equivalent original operation.
		}
\end{figure}

The physical dimensions of each of the two the synapse arrays on the hardware is fixed to $N = 256$ rows ($128$ for signed weights) and $M = 256$ columns limiting the shape of a single multiply-accumulate (MAC) operation.
Temporal reuse of the synapse array allows larger operations.
The operation is split into parts which fit on a single synapse array and are placed individually with a round-robin allocation scheme on the available synapse arrays.
\Cref{fig_partitioning} visualizes the partitioning scheme.
In order to support more columns, the results of the split operations are to be concatenated.
To the contrary, to support more rows the results of the split operations are to be summed up digitally:
\begin{equation}
y_j = \sum_i^N x_i w_{ij} = \left(\sum_i^{N_1} x_i w_{ij}\right) + ... + \left(\sum_i^{N_R} x_i w_{ij}\right),\quad N = \sum_r^R N_r
\end{equation}
where the input size $N$ is split into $R$ ranges $N_r$ of analog computation $\sum_i^{N_r} x_i w_{ij}$, which are then summed digitally.
This approach is expected to be comparable to computation on a larger physical synapse array if no boundary effects ---like analog saturation or digital overflow--- occur.

For weight matrices large in both dimensions compared to the synapse array this partitioning scheme leads to a optimal chip area usage, because the number of partial synapse array allocations scales with the edges like $\mathcal{O}(N + M)$ while the number of full synapse array allocations scales with the area like $\mathcal{O}(N \cdot M)$.

\pagebreak[4]
\subsection{Parallel execution of convolutional layers}\label{sec:expansion}
\begin{wrapfigure}[12]{r}{0.26\textwidth}
\vspace{-2\baselineskip} %
\scalebox{0.784}{\begin{tikzpicture}[node distance=1.2cm,
    every node/.style={font=\sffamily}, align=center,>={Latex[width=2mm,length=2mm]}]

	\draw[draw, fill=color2!40] (0,0) rectangle ++(0.35,3);
	\node (x) at (0.175,3-0.175) {x};

	\draw[draw, fill=white] (0.5,0) rectangle ++(3.15,3);

	\foreach \p in {0,...,5} {
		\draw[draw, densely dashed, fill=color4!40] (0.5 + \p * 0.525, 3 - \p * 0.3) rectangle ++(0.525, -1.5);
	}
	\node (w) at (0.5 + 0.2625,3-0.175) {w};

	\draw[draw, fill=color3!40] (0.5,-0.5) rectangle ++(3.15,0.35);
	\node (y) at (0.5 + 0.175,-0.5+0.175) {y};

	\begin{scope}[on background layer]
		\draw[draw=none, fill=color0!20] (-0.15, -0.65) rectangle ++(3.95,3.8);
	\end{scope}
\end{tikzpicture}}
\vspace{-1.5\baselineskip}
\caption{Expanded \texttt{Conv1d} layer maximizing usage of the synapse array.}\label{fig:expansion}
\end{wrapfigure}
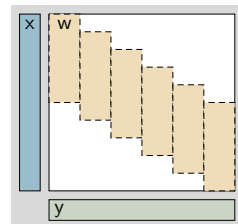%
In convolutional layers, the size of the transformed weight matrix (cf. \cref{fig_conv}) often is a lot smaller than the synapse array of our accelerator.
Especially in the one dimensional case, it is possible to perform multiple such operations in one step, a possible layout on the chip is sketched in fig.~\ref{fig:expansion}.
The downside of this approach is the increase of independent parameters, since the weights of each expansion have to be learned individually due to fixed pattern noise and other deviations, cf.~\cite{weis2020inference}.
To overcome this problem, we add modified versions of the data set to the training data, shifted by the stride of the convolution operation.
However, this is only applicable for a single layer or an equal number of parallel executions, since it can only be tuned to the hyper-parameters of one layer. 
Nevertheless, the execution time while evaluating the model is reduced up to a factor of the number of parallel executions and the energy consumption is nearly decreased by the same factor.

\subsection{Handling hardware setup, initialization and parameters}
The accelerator hardware setup and initialization routine is time-consuming compared to a single computation.
Hence, initialization of the hardware is only performed once.
We also allow users to modify the hardware initialization process.
Exclusive access to hardware resources is handled via free functions utilizing a singleton pattern.
Inside the execution of an operation, the JIT executor, cf.~\cref{section_execution}, then uses available hardware acquired previously via the singleton.

Using \BSS{2} involves choice of parameters affecting the available dynamic range.
Two such parameters, described in~\cite{weis2020inference}, are the interval duration between successive input events, and sending the same input multiple times.
Both parameters were introduced to increase the precision of the analog computation on first-version chips.
As such parameters possibly have to be tuned for each layer we supply them side-by-side to other operation parameters already present in the equivalent CPU operation.
The following listing shows the implemented scheme at the example of a \texttt{conv1d} operation:
\begin{center} %
\texttt{conv1d(x, kernel, stride=1\textcolor{color2}{, num\_sends=6, wait\_between\_events=25})}
\end{center}
While these parameters are in principle differentiable, a model is yet to be developed.
Therefore they are treated as non-differentiable hyperparameters.

Since the hardware specific parameters are not present for CPU/GPU operations, care has to be taken when converting such operations to accelerator hardware execution.
We provide replacements for the PyTorch layers \texttt{Linear} and \texttt{Conv\{1,2\}d}, which take these additional parameters as keyword arguments.
Since these layers use the same state as their counterparts, pre-trained weights can easily be loaded into a model that uses them.

\section{Results}\label{sec:results}

We first look at performance figures for speed of execution and utilization of the accelerator hardware in order to evaluate imposed software overhead.
Afterwards, we demonstrate usability of the presented software framework by an application on the human activity recognition dataset~\cite{anguita2013public}.

\subsection{Performance Evaluation}\label{section_performance}

\paragraph{Hardware Limitations and Measurement Setup}
The currently used first hardware version (v1) contains a bug which requires rewriting the synapse array for each sent input.
As a consequence, the input data volume increases by a factor of $\approx 100$.
To increase the precision of the analog calculations, inputs are repeated.
In addition, successive events are spaced over time, cf.~\cite{weis2020inference}.
The second hardware version (v2) is currently in the commissioning process and expected to be free of these limitations.
Performance estimations for v2 are given by disabling these workarounds on v1.
This only affects the quality of the computation.

The available hardware setup consists of the \BSS{2} chip connected to an FPGA which provides one GbE link to the host computer.
This connection poses a severe communication limitation as the chip features full-duplex \SI{8}{\giga\bit\per\second} interconnects.
To evaluate software performance against the chip hardware design limitation, we use a software simulation providing a fast mock-up communication partner.
This is a valid approach for software performance evaluation since the postprocessing of response data is content-agnostic.

\paragraph{Results}

\begin{figure}[tb]
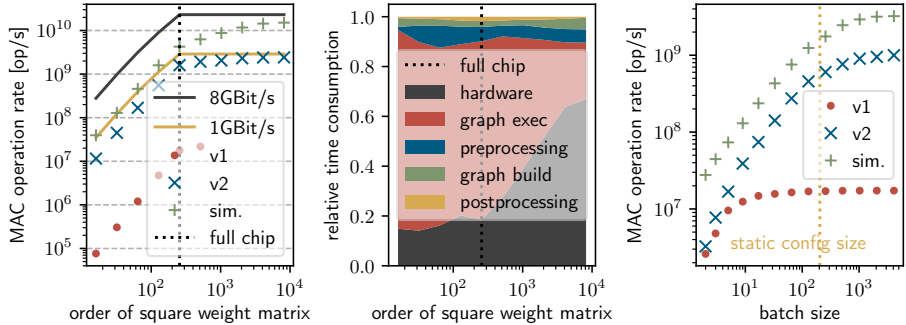

	\centering
	\hspace{-.60em}
	\scalebox{0.74}{\input{plots/perf_quadratic.pgf}}%
	\hspace{-.60em}%
	\scalebox{0.74}{\input{plots/perf_batch_size.pgf}}
	\vspace{-\baselineskip}
	\caption{%
	\label{fig_results_quadratic_growth}%
	Performance measurement for square matrices with sizes ranging from $1^2$ to $(2^{14})^2$ elements (batch size: $2000$).
	The dotted line marks the matrix size of a single \BSS{2} chip (left and center).
	\emph{Left}: Rate of MAC operations for real and simulated hardware:
	the yellow line illustrates the limiting speed for setups using \SI{1}{\giga\bit\per\second} links (host link);
	the gray line shows this for \SI{8}{\giga\bit\per\second} links (chip I/O);
	disregarding a constant configuration overhead, the rate of operations increases linearly for matrices smaller than a chip;
	for matrices larger than the full chip area, individual runs can overlap in the pre- and postprocessing step thereby increasing the speed further;
	the red dots and blue crosses show measurements using the real hardware setup.
	\emph{Center}: Distribution of execution times for the \SI{8}{\giga\bit\per\second} case (cf.~\cref{fig_matmul} for details on the categories);
	the chip utilization increases for matrices larger than the full chip area.
	\label{fig_results_batch_size}%
	\emph{Right}:
	Measurement for varying batch sizes (matrix size: $256^2$);
	rates are plotted for different chip versions and simulated design-goal hardware;
	the dotted line marks the batch size where the static configuration overhead matches the variable data volume;
	this coincides with the point where $\approx 50\%$ of the maximum performance is reached.
	}
\end{figure}

We evaluate the performance in terms of multiply-accumulate (MAC) operations per time for square-shaped weight matrices with fixed batch size and varying batch size for a fixed weight matrix in \cref{fig_results_batch_size}.
The left panel shows that the current implementation is able to saturate v2 in combination with a 1-GbE host connection in the limit of large matrix multiplications.
Furthermore, the current implementation reaches up to the hardware design limitation within a factor of two.
Given the static configuration necessary for a matrix multiplication, the right panel shows the achieved performance for v2 and the design limitation simulation is within $50\%$ of the saturation speed for batch sizes larger than $\approx 200$.

\subsection{Application example: Human Activity Recognition}

Our application example uses the smartphone acceleration data from \cite{anguita2013public}.
The dataset contains recordings of 30 subjects carrying a waist-mounted smartphone while walking straight ahead, up or down, sitting, standing or lying on their backs.
The data is already split in training and test data, containing 9 channels of sensor data with a length of \SI{2.56}{s} sampled at \SI{50}{Hz}, each.

\paragraph{The model}
We use a 1-d convolution layer with kernel size 32, stride 6 for feature detection, followed by two dense layers.
The model topology is defined by \cref{tab:har_model_overview}.

\begin{table}[tb]
	\centering
    \caption{\label{tab:har_model_overview}%
		Model parameters for ``Human Activity Recognition'' example.
	}
    \begin{tabular}{l@{\hskip.4cm}l@{\hskip.4cm}l@{\hskip.4cm}l@{\hskip.4cm}l}
     Layer & Activation & Input Shape & Output Shape & \# of Params \\
     \toprule
             Conv1d   & ReLU    & [-1, 9, 128] & [-1, 16, 16] & \hphantom{32'000}\llap{4'608} \\
             Linear-1 & ReLU    & [-1, 256]    & [-1, 125]    & \hphantom{32'000}\llap{32'000} \\
             Linear-2 & Softmax & [-1, 125]    & [-1, 6]      & \hphantom{32'000}\llap{750}
    \end{tabular}
\end{table}

\noindent
The hyper-parameters of all layers are optimized for the dimensions of the analog substrate to achieve a balance between accuracy, energy efficiency and execution speed.
We do not use additive biases in any layer.

\begin{figure}[tb]
    \centering
	\vspace{-1\baselineskip}
    \scalebox{0.66}{\input{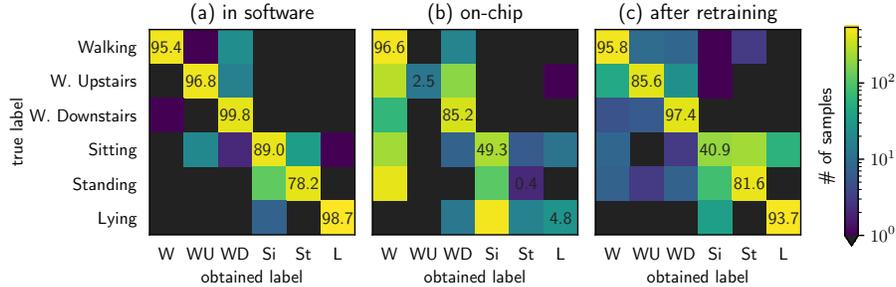}}
	\vspace{-1.5\baselineskip}
    \caption{\label{fig:conf_matrix_har}%
                Confusion matrices and recall accuracies for the separate test set, (a) after 50 epochs training in software, (b) executed on \BSS{2} without retraining, and (c) after one additional epoch training with hardware in the loop.
}
\end{figure}

\paragraph{Training}
First, we trained the model for 50 epochs in software without our accelerator.
In this step, the weights and inputs were already quantized and scaled to the dynamic range of the hardware, and Gaussian noise was added to the output of each layer.
The resulting confusion matrix is shown in \cref{fig:conf_matrix_har}a.
Running the very same model on our analog substrate shows results that are quite off (cf.~\cref{fig:conf_matrix_har}b), which can be explained with fixed pattern noise and non-linearities.
One additional epoch with hardware-in-the-loop training is sufficient to come significantly closer to our software results as is depicted in \cref{fig:conf_matrix_har}c.

\paragraph{Results}
The achieved overall accuracy is 92.7\% in software and 82.3\% on-chip.
As already found in \cite{anguita2013public,ronao2016human_shorter} there is a noticeable misclassification between sitting and standing.
This can be explained by the similar orientation of the smartphone and the non-dynamic nature of these activities.

\section{Discussion and Outlook}\label{sec:discussion}

This work presents software integrating analog vector-matrix multiplication on \BSS{2} into the PyTorch framework.
Using the PyTorch extension interface, \texttt{hxtorch} provides support for convolutional and dense layers.
The hardware backend builds upon the BrainScaleS operating system \cite{mueller2020bss1_nourl,mueller2020bss2ll_nourl} and utilizes a configuration and runtime flow that is essentially identical to the spiking operation of the system.
We describe the underlying implementation comprising data flow representation, operator mapping, hardware partitioning, system setup and execution.
We evaluate end-to-end runtime performance and show that the best possible performance, in terms of hardware design limitations, is reached within a factor of two for sufficiently large matrix multiplications.
We obtain \SI{14.7}{\giga\operation\per\second} for a single simulated accelerator.
In conjunction with the single GbE host link, this surpasses the speed of v2 by a factor of $\approx 5$.
The static configuration of a $256$-by-$256$ matrix matches a batch size of around $200$ $256$-wide inputs in terms of data volume and computation time.
Finally, we demonstrate an end-to-end application example on ``Human Activity Recognition'' with software results comparable to~\cite{ronao2016human_shorter}
but obtain a drop in the classification quality for the hardware backend.
We expect the result to improve for v2.

\label{sec:outlook}%
Although we focused here on an interface to the non-spiking functionality of the accelerator, it is possible to extend PyTorch to model the spiking operation of the chip as well.
In the simplest case input spikes are represented as sparse binary tensors, with one of the axes being the time dimension.
For efficient hardware operation this time dimension is identified with the time dimension of the intrinsic temporal dynamics of the analog circuitry.
This implies that recurrent spiking neural networks with up to $N = 512$ neurons per chip can be implemented.
A simulation can be run concurrently to estimate the gradients.
By performing the forward pass through chip and backward pass through the simulation like in \cite{legallo2018mixedprecision_nourl,schmitt2017hwitl_nourl,cramer2020training_nourl}, hardware parameters can be adjusted based on simulation parameter updates.
Ideally, gradient estimation would be independent of analog measurements on the chip, which introduce additional memory bandwidth requirements.
Finally, deeper integration into ML frameworks facilitates compute graph analysis.
Consequently, network topologies can be described natively;
the compute graph still allows for the specification of arbitrary computation ---e.g., plasticity rules in spiking neural networks--- that can be offloaded to the embedded microprocessors.

\section*{Contributions and Acknowledgments}

PS is the main developer of the software extensions of this work. %
EM is the lead developer and architect of the \BSS{2} software stack.
AE contributed to the PyTorch extension module and the model application.
AE, AL, CM, OB, TCW and YS contributed to the software implementation.
CP contributed to the initial implementation of the PyTorch extension module.
JW is a main contributor to hardware commissioning and contributed to the PyTorch extension module.
SB contributed to the hardware design, commissioning and the software implementation.
SS contributed to core software components and the software design.
JS is the lead designer and architect of the \BSS{2} neuromorphic system.
All authors discussed and contributed to the manuscript.

The authors wish to thank all present and former members of the Electronic Vision(s) research group contributing to the BrainScaleS-2 hardware platform. %
The authors express their special gratitude towards:
J.\ Klähn, D.\ Stöckel and S.\ Friedmann for earlier software contributions.
We especially express our gratefulness to the late Karlheinz Meier who initiated and led the project for most if its time.
This work has received funding from the EU (%
[H2020/2014-2020]%
)
under grant agreements
720270, %
785907, %
945539 (HBP) %
as well as from the BMBF~(16ES1127).%

\printbibliography[notkeyword=own_software]
\printbibliography[title={BrainScaleS Software Repositories},keyword=own_software]

\end{document}